%% file: main_arXiv.tex
\documentclass[letterpaper]{article} 
\usepackage{aaai23}  
\usepackage{times}  
\usepackage{helvet}  
\usepackage{courier}  
\usepackage[hyphens]{url}  
\usepackage{graphicx} 
\usepackage{natbib}  
\usepackage{caption} 
\usepackage{algorithm}
\usepackage{algorithmic}

\usepackage{newfloat}
\usepackage{listings}
\DeclareCaptionStyle{ruled}{labelfont=normalfont,labelsep=colon,strut=off} 
\floatstyle{ruled}
\newfloat{listing}{tb}{lst}{}
\floatname{listing}{Listing}
\usepackage{xcolor}
\usepackage{adjustbox}
\usepackage{multirow}
\usepackage{subcaption}
\usepackage{amsmath}
\usepackage{makecell} 
\usepackage{lipsum}
\usepackage{amssymb}

\makeatletter
\usepackage[symbol]{footmisc}

\title{DBN-Mix: Training dual-branch network using bilateral mixup augmentation for long-tailed visual recognition}
\author{Jae Soon Baik, In Young Yoon, and Jun Won Choi\thanks{Corresponding Author}
}
\affiliations{
    Department of Electrical Engineering, Hanyang University

    \{jsbaik, inyoungyoon\}@spa.hanyang.ac.kr, junwchoi@hanyang.ac.kr
}

\usepackage{bibentry}

\begin{document}

\maketitle

\begin{abstract}
There is growing interest in the challenging visual perception task of learning from long-tailed class distributions. The extreme class imbalance in the training dataset biases the model to prefer recognizing majority class data over minority class data. Furthermore, the lack of diversity in minority class samples makes it difficult to find a good representation. In this paper, we propose an effective data augmentation method, referred to as {\it bilateral mixup augmentation}, which can improve the performance of long-tailed visual recognition. The bilateral mixup augmentation combines two samples generated by a uniform sampler and a re-balanced sampler and augments the training dataset to enhance the representation learning for minority classes. We also reduce the classifier bias using class-wise temperature scaling, which scales the logits differently per class in the training phase. We apply both ideas to the {\it dual-branch network (DBN)} framework, presenting a new model, named {\it dual-branch network with bilateral mixup (DBN-Mix)}. Experiments on popular long-tailed visual recognition datasets show that DBN-Mix improves performance significantly over baseline and that the proposed method achieves state-of-the-art performance in some categories of benchmarks.
\end{abstract}

\section{Introduction}
\label{sec:introduction}
Deep neural networks (DNNs) have achieved great success in a variety of visual perception tasks thanks to publicly available large datasets such as ImageNet \cite{deng2009imagenet} and MS COCO \cite{lin2014microsoft}. Although the classes of images in these recognition datasets are balanced to have an approximately uniform distribution, large-scale real-world datasets have a long-tailed distribution; few classes occupy most of the data, while most classes have few samples. Standard supervised learning on a long-tailed dataset tends to be severely biased toward majority classes, resulting in poor classification accuracy for minority classes. This trend is problematic in applications such as autonomous driving, where image recognition of all other classes is equally important. They raise the challenge of designing effective training methods for {\it long-tailed datasets}, which can improve the recognition performance for both majority and minority classes.

\begin{figure*}[t]
    \centering
    \begin{subfigure}[]{0.32\textwidth} 
        \includegraphics[width=1.0\textwidth]{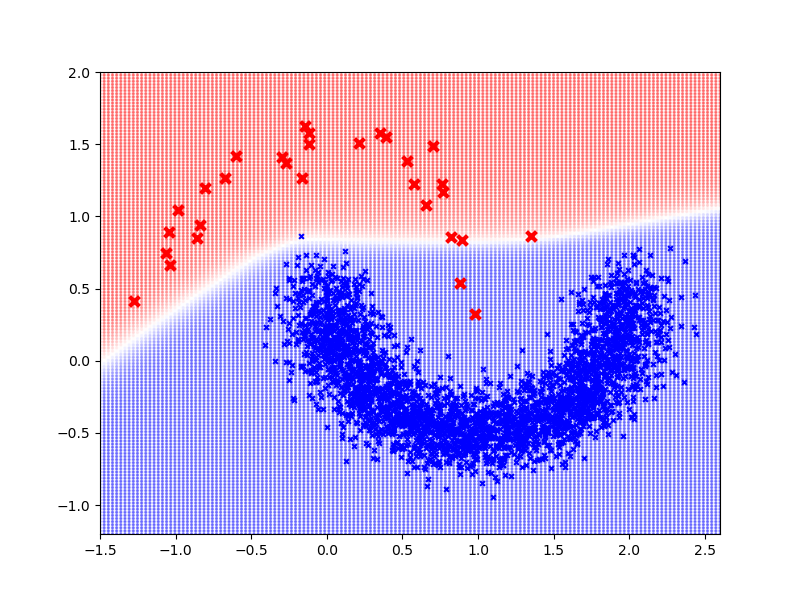}
        \caption{ERM}
    \end{subfigure}
    \centering
    \begin{subfigure}[]{0.32\textwidth} 
        \includegraphics[width=1.0\textwidth]{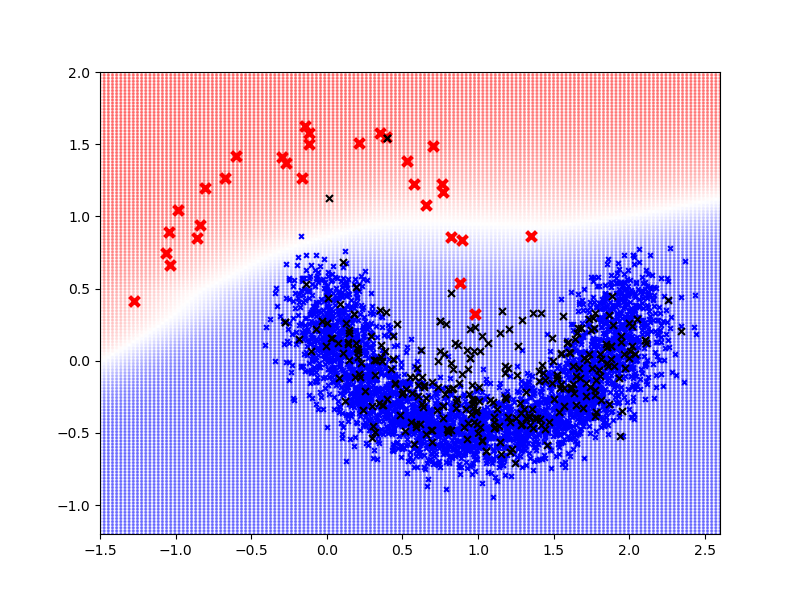}
        \caption{Mixup}
    \end{subfigure}
    \centering
    \hspace{0.017\textwidth} 
    \begin{subfigure}[]{0.32\textwidth} 
        \includegraphics[width=1.0\textwidth]{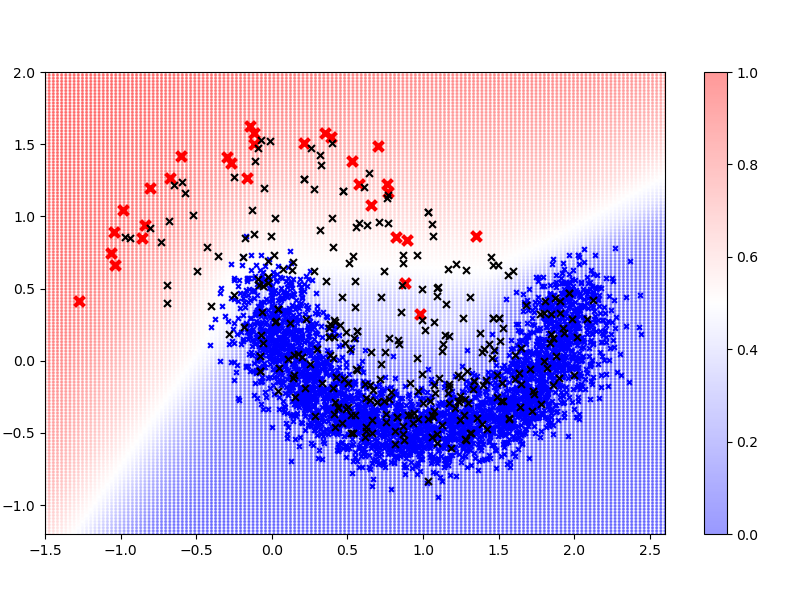}
        \caption{Bilateral Mixup}
    \end{subfigure}
    \caption{Outputs of the classifiers trained by (a) empirical risk minimization (ERM), (b) mixup, and (c) bilateral mixup: Three-layer neural networks were trained on the imbalanced two half-moon dataset with the imbalance ratio of 100. Red and blue cross marks correspond to minority class samples and majority class samples, respectively. The black cross marks  indicate the training samples generated by the data augmentation.} 
    \label{fig:bilateral_mixup}
\end{figure*}

Various training methods for long-tailed recognition tasks have been proposed to date \cite{cao2019ldam, cui2019class, wang2021ride, zhong2021mislas}. Multiple expert networks have been used to address class imbalance issues, where multiple models are jointly trained to model both majority and minority class data \cite{xiang2020lfme, zhou2020bbn, wang2021ride}. BBN \cite{zhou2020bbn} employed a dual-branch network (DBN) consisting of two parallel branches called a {\it conventional learning branch} and a {\it re-balancing branch}. The conventional learning branch was trained using a {\it uniform sampler} while the re-balancing branch was trained using a {\it re-balanced sampler}, which generated the samples with inversely proportional class distribution to the original dataset. The re-balancing branch can effectively alleviate the classifier bias, but it tends to oversample minority class samples, thereby degrading the quality of the representation.

An alternative method for addressing the class imbalance issue is data augmentation \cite{chawla2002smote, li2021metasaug, chu2020feature, kim2020m2m, zhang2021bag, chou2020remix, park2022cmo}. 
This method is particularly useful for long-tailed datasets, as synthetically generated samples can mitigate the lack of minority class data. Various data augmentation strategies have been proposed for long-tailed recognition task \cite{li2021metasaug, chu2020feature, kim2020m2m, zhang2021bag, chou2020remix, park2022cmo}. However, due to its limited size and diversity, it is difficult to generate samples that follow the true distribution of minority class data.

In this paper, we present a simple yet effective data augmentation strategy, referred to as {\it bilateral mixup augmentation}, designed to address the class imbalance for long-tailed recognition tasks. The proposed bilateral mixup differs from conventional mixup \cite{Zhang2018mixup} in that a {\it class distribution-aware mixup strategy} is used to combine the samples from a uniform sampler and a re-balanced sampler. The samples generated by the proposed mixup operations are located near the boundaries of minority class regions, where data points are sparsely distributed, and serve to better capture the distribution of minority classes as distinct from other classes. Our class distribution-aware combination rule significantly enhances the ability of the original mixup augmentation to improve data representation, especially for long-tailed class distributions. Fig. \ref{fig:bilateral_mixup} shows a toy example that demonstrates the effect of the proposed bilateral mixup as compared to the original mixup. Without the proposed combination rule, minority class samples would not participate in sample generation, failing to improve classification performance. On the other hand, the proposed mixup method produces a decision boundary that better separates the minority class points from the other samples. 

While bilateral mixup improves the representation ability, we also need a measure to compensate for the bias of the classifier. We present a {\it class-wise temperature scaling} method that applies class-dependent temperature parameters to the logits of the classifier. In the training phase, the model is trained with this logit scaling enabled to have higher margins for minority classes. The logit scaling is disabled during the inference phase. 

We integrate the above two ideas into the multiple expert network framework, presenting a new architecture, the so-called {\it dual-branch network with bilateral mixup (DBN-Mix)}. We extensively evaluated the performance of the proposed DBN-Mix on widely used long-tailed visual recognition datasets: CIFAR-LT-10, CIFAR-LT-100, ImageNet-LT, and iNaturalist 2018. The proposed DBN-Mix significantly outperforms conventional training methods designed for long-tailed visual recognition and achieves state-of-the-art performance in some categories of benchmarks.

The main contributions of this study are summarized as follows:

\begin{itemize}
    \item We present a simple yet effective data augmentation method designed to improve long-tailed visual recognition performance. We propose a novel mixup operation that combines two samples drawn from the dataset with different sampling distributions. The class distribution-aware mixup strategy serves to better model minority classes, improving classification accuracy.
    \item Our study addresses two sources of performance degradation caused by long-tailed class distributions: 1) poor representation of minority class data due to lack of diversity in samples and 2) classifier bias caused by imbalanced class distribution. 
    Our analyses show that the proposed bilateral mixup and class-wise temperature scaling effectively mitigate the performance degradation in both representation learning and classifier learning. While numerous existing methods have used two-stage training to improve both representation learning and classifier learning \cite{kang2020decouple, wang2017learning, tang2020deconfound, zhang2021bag, li2022gcl}, our class-dependent mixup method enables {\it end-to-end training}, which results in improved performance. 
    \item The proposed method can be implemented only with a few lines in the code and does not require a complex optimization process like other data augmentation methods \cite{zhang2021bag, kim2020m2m}. The proposed ideas are also adaptable, as they can be applied to any network architecture for the long-tailed visual recognition task. We demonstrate that our ideas can also be simply integrated into the common single-branch network (SBN) and achieve significant performance improvements over the baseline, although its classification accuracy is not as high as that achieved by DBN-Mix.  
\end{itemize}

\section{Related Work}
\label{sec:related work}
\subsection{Re-sampling and Re-weighting.}
Re-sampling and re-weighting methods have been extensively researched. The re-sampling strategy balances the class distribution by oversampling minority class data \cite{chawla2002smote, han2005borderline, buda2018systematic, byrd2019effect, shen2016relay} or undersampling majority class data \cite{buda2018systematic, drummond2003c4}. Re-weighting modifies the loss function based on class- or sample-level criteria \cite{cao2019ldam, cui2019class, lin2017focal, shu2019metaweight, ren2018learning}. Cao $et\ al.$ \cite{cao2019ldam} proposed a label-distribution-aware margin loss based on the theoretical margin bound, and Cui $et\ al.$ \cite{cui2019class} introduced the notion of an effective number to determine re-weighting factors. Several studies have employed meta-learning to determine a strategy for weighting the loss function \cite{shu2019metaweight, muhammad2020metaclass, ren2018learning}. 

\subsection{Two-stage Training Strategy.} 
Several recent studies have investigated the impact of long-tailed class distribution on representation learning and classifier learning and have proposed two-stage training methods to improve both  \cite{kang2020decouple, wang2017learning, tang2020deconfound, zhang2021bag, zhong2021mislas, li2022gcl}. The pioneering work in \cite{kang2020decouple} first used normal training data to train the backbone network and then used class-balanced samples to refine the classifier only. Since then, several two-stage training methods, including the CAM-based method \cite{zhang2021bag}, logit adjustment loss \cite{menon2021logitadjust}, MetaSAug \cite{li2021metasaug}, and MiSLAS \cite{zhong2021mislas}, have been proposed. Although our method attempts to solve the class imbalance issue from both representation learning and classifier learning perspectives, the proposed DBN-Mix does not require two-stage training and allows end-to-end learning. 

\begin{figure*}[t] 
    \centering
    \includegraphics[width=0.85\textwidth]{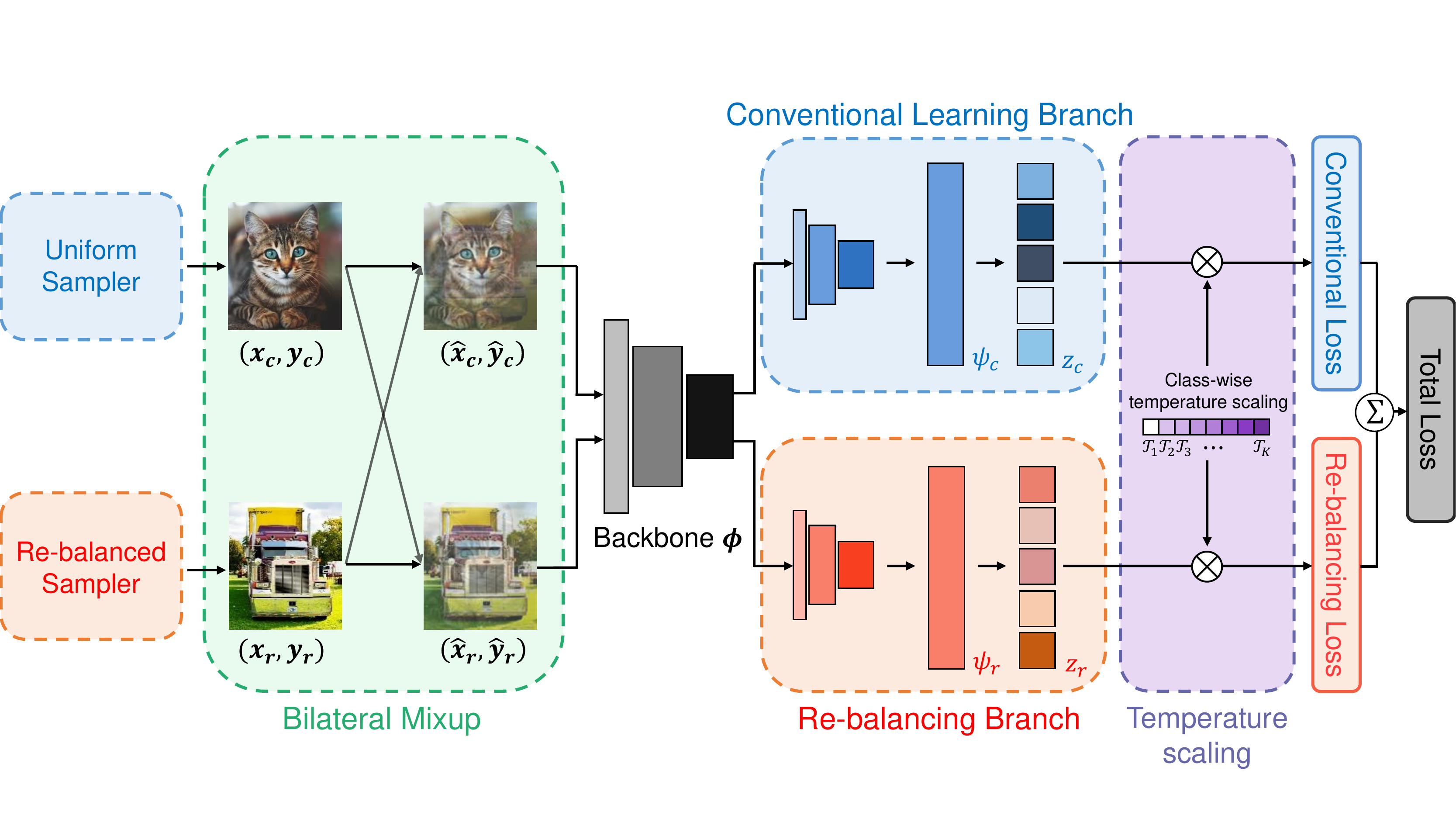}
    \caption{{\bf Overview of the proposed method:} Our method use two bilateral mixup samples $(\hat{x}_c, \hat{y}_c)$ and $(\hat{x}_r, \hat{y}_r)$ to train two branch networks, conventional learning branch and re-balancing branch. In the training phase, we use $(\hat{x}_c, \hat{y}_c)$ for conventional learning branch and $(\hat{x}_r, \hat{y}_r)$ for re-balancing branch. For the inference phase, two prediction logits from each branch are averaged to return the final output.}
    \label{fig:overall_framwork}
\end{figure*}

\subsection{Ensemble-based Approach} 
Ensemble-based methods for long-tailed visual recognition have been actively studied \cite{zhou2020bbn, xiang2020lfme, wang2021ride}. These methods use multiple ensemble models to model data with different class distributions. BBN \cite{zhou2020bbn} employed two branch networks, where the conventional learning branch was trained by the uniform sampler and the re-balancing branch was trained by the re-balanced sampler. LFME \cite{xiang2020lfme} trained multiple networks on subsets of the dataset and then aggregated the information from the subnetworks using knowledge distillation. RIDE \cite{wang2021ride} employed the multiple networks to reduce both the model bias and variance using distribution-aware loss. In \cite{guo2021collaborative}, the cross-branch consistency loss between two branch networks was used as a regularizer for multi-label visual recognition learning.

\subsection{Data Augmentation.}
A data augmentation method has been used to handle the data scarcity of minority classes  \cite{Zhang2018mixup, zhang2021bag, li2021metasaug, chu2020feature, kim2020m2m, chou2020remix, park2022cmo, chawla2002smote}. As mixup augmentation has been proposed to improve the generalization of the model \cite{Zhang2018mixup}, it has been adapted to solve the problem of long-tailed visual perception \cite{chou2020remix, zhong2021mislas, zhang2021bag}. Remix \cite{chou2020remix} employed the mixup strategy that assigned higher mixing factors to labels associated with minority class samples. MiSLAS \cite{zhong2021mislas} used the mixup only for the first-stage of training and then used label-aware smoothing to deal with classifier bias. CMO \cite{park2022cmo} proposed a data augmentation method based on CutMix that can transfer rich contexts from majority to minority samples. Several methods \cite{kim2020m2m, li2021metasaug} attempted to use the knowledge transfer to generate synthetic minority class samples using the learned representation of majority classes. However, these methods require a carefully designed optimization process to achieve their goals.

\section{Proposed Method}
\label{sec:proposed method}
In this section, we present the details of the proposed DBN-Mix method. 

\subsection{Overview of DBN-Mix}
\label{sec:overall}
Consider a $K$-class image classification task. We train a backbone network $\phi$ and a classifier network $\psi$ in an end-to-end fashion on the long-tailed training set. Let $\mathcal{D}=(X, Y)$ be the training dataset, where $X=\{x_1, ..., x_N\}$, $Y=\{y_1, ..., y_N\}$ and $x_i$ and $y_i$ are the $i$th image sample and the corresponding label, respectively. $N$ denotes the cardinality of the training dataset. The label $y_i$ is encoded by a one-hot vector $[y_{i,1}, ..., y_{i, K}]^T\in \{0, 1\}^K$. 

Fig. \ref{fig:overall_framwork} depicts the structure of DBN-Mix. It consists of the shared backbone network $\phi$ followed by two branch subnetworks, the conventional learning branch $\psi_{c}$ and the re-balancing branch $\psi_{r}$. Two separate mini-batches are constructed using a uniform sampler and a re-balanced sampler to train each branch network. 
The uniform sampler draws a sample with an equal probability $P=1/N$, where $N$ is the cardinality of the training set. The re-balanced sampler draws a sample from the class $k$ with a probability  
\begin{align}
    P_k &= \frac{w_k}{\sum_{k=1}^K w_k}, 
    \label{eq:prop_r} \\
    w_k&=\left(\frac{N_{max}}{N_k}\right)^{\frac{1}{\gamma}}, 
    \label{eq:frequency gamma}
\end{align}
where $\gamma$ is the hyperparameter, $N_k$ is the sample size of class $k$, and $N_{max}$ is the maximum sample size for all the classes. Hyperparameter $\gamma$ adjusts the reversed class distribution. As $\gamma$ increases, the re-balancing effect increases. When $\gamma$ goes to infinity, each class is chosen with an equal probability. 
Two samples  $x_c$ and $x_r$ are generated by the uniform and re-balanced samplers, respectively, and then are transformed to $\hat{x}_c$ and $\hat{x}_r$ by the proposed bilateral mixup operation. The transformed samples are then fed to the shared backbone network followed by two subsequent branch subnetworks 
\begin{align}
    z_c &= \psi_c (\phi(\hat{x}_c)), \\
    z_r &= \psi_r (\phi(\hat{x}_r)),
    \label{eq:logits}
\end{align}
where $z_c=[z_{c,1}, ..., z_{c, K}]^T$ and $z_r=[z_{r,1}, ..., z_{r, K}]^T$ are the $K$-dimensional logits from the two subnetworks. 

\subsection{Bilateral Mixup Augmentation}
\label{sec:bilateral mixup augmentation}
Recall that the original mixup augmentation \cite{Zhang2018mixup} generates the sample $(\tilde{x}, \tilde{y})$ by taking a convex combination of two samples $(x_i, y_i)$ and $(x_j, y_j)$, i.e., 
\begin{align}
    \lambda &\sim Beta(\alpha, \alpha), \\
    \left( \begin{matrix} \tilde{x} \\ \tilde{y} \end{matrix} \right) &= \lambda 
\left(\begin{matrix} x_i \\  y_i \end{matrix}\right) + (1-\lambda) \left(\begin{matrix}x_j \\ y_j \end{matrix} \right), \label{eq:mixup}
\end{align}
where $Beta(\cdot, \cdot)$ denotes a beta distribution and $\alpha$ denotes the hyperparameter for the beta distribution. The bilateral mixup augmentation simply takes two samples $({x}_c, {y}_c)$ and $({x}_r, {y}_r)$ from the uniform sampler and the re-balanced sampler, respectively, and combines them with different ratios, i.e.,  
\begin{align}
    \lambda &\sim Beta(\alpha, \alpha), \label{eq:lambda}\\
    \lambda_c\; &= \max(\lambda, 1-\lambda), \label{eq:mixup_lambda_c}\\
    \lambda_r\; &= \min(\lambda, 1-\lambda), \label{eq:mixup_lambda_r}\\
    \left(\begin{matrix} \hat{x}_c \\ \hat{y}_c \end{matrix}\right) &= \lambda_c \left(\begin{matrix} {x}_c \\ {y}_c \end{matrix}\right) + (1-\lambda_c) \left(\begin{matrix} {x}_r \\ {y}_r \end{matrix}\right), \label{eq:mixup_input_c}\\
          \left(\begin{matrix} \hat{x}_r \\ \hat{y}_r \end{matrix}\right) &= \lambda_r \left(\begin{matrix} {x}_c \\ {y}_c \end{matrix}\right) + (1-\lambda_r) \left(\begin{matrix} {x}_r \\ {y}_r \end{matrix}\right). \label{eq:mixup_input_r}
  \end{align}
Two bilateral mixup samples $(\hat{x}_c, \hat{y}_c)$ and $(\hat{x}_r, \hat{y}_r)$ are fed to the conventional learning branch and the re-balancing branch, respectively.



\subsection{Class-wise Temperature Scaling}
\label{sec:temperature}
During the training period, class-wise temperature scaling is applied to the logits $z_c$ and $z_r$ 
\begin{align}
    \hat{p}_{c, k}&=\exp\left(\frac{z_{c, k}}{\mathcal{T}_k}\right)/\sum_{k=1}^K\exp\left(\frac{z_{c, k}}{\mathcal{T}_k}\right), \\
    \hat{p}_{r, k}&=\exp\left(\frac{z_{r, k}}{\mathcal{T}_k}\right)/\sum_{k=1}^K\exp \left(\frac{z_{r, k}}{\mathcal{T}_k}\right),
    \label{eq:scaled prediction}
\end{align}
where $\mathcal{T}_k$ denotes the temperature parameter. The parameter $\mathcal{T}_k$ is set differently per class to reduce the classifier bias caused by the long-tailed class distribution. The parameter $\mathcal{T}_k$ is given by 
 \begin{align}
   \mathcal{T}_k &= \left(\frac{\max(\mathcal{B}_{1:K})}{\mathcal{B}_k}\right)^{\frac{1}{\eta}}, \label{eq:temperature} \\
    \mathcal{B}_k &= \epsilon \frac{N_k}{N_{max}} + (1-\epsilon),
    \label{eq:temperature b}
\end{align}
where $\eta$ and $\epsilon$ are hyperparameters, $\mathcal{B}_{1:K}=\{\mathcal{B}_{1},..., \mathcal{B}_{K}\}$,  $N_k$ is the number of samples in the $k$th class and $N_{max}$ is the maximum sample size for all classes. 
This leads to a lower temperature value $\mathcal{T}_k$ for the majority classes and a higher temperature value $\mathcal{T}_k$ for the minority classes.
During training, class-wise temperature scaling encourages the model to favor minority class samples over majority class samples. Prior studies proposed a bias-based logit adjustment method to compensate for the classifier bias \cite{cao2019ldam, cui2019class, menon2021logitadjust}. These methods are not suitable for use with the proposed bilateral mixup because the labels $\hat{y}_c$ and $\hat{y}_r$ have soft values due to the mixup operation. Class-wise temperature scaling allows for accurate control of classifier bias throughout the training phase.

The loss function used to train the DBN-Mix is composed of the following terms  
\begin{align}
    \mathcal{L}_{total} &= \frac{1}{2}\mathcal{L}(\hat{p}_c, \hat{y}_c) + \frac{1}{2}\mathcal{L}(\hat{p}_r, \hat{y}_r),
    \label{eq:total loss}
\end{align}
where $\hat{p}_c = [\hat{p}_{c,1},...,\hat{p}_{c,K}]^T$, $\hat{p}_r = [\hat{p}_{r,1},...,\hat{p}_{r,K}]^T$, and $\mathcal{L}$ denotes the cross-entropy loss, $\mathcal{L}(p, y) = - \sum_{k=1}^{K} y_{k} \log p_{k}$. 
Note that the entire network is trained in an end-to-end manner. 

\subsection{Model Inference}
During the inference phase,  a single test image $x$ is fed into the two branch networks. The outputs from the dual-branch subnetworks are then combined using equal weights
\begin{align}
    z &= \frac{1}{2}(\psi_c (\phi(x)) + \psi_r (\phi(x))).
    \label{eq:final prediction}
\end{align}
Finally, the softmax function is applied to the combined logit $z$ without class-wise temperature scaling. 

\subsection{Application to Single-Branch Network}
While bilateral mixup is primarily intended for DBN architecture, it can also be applied to SBN with a common single-branch structure. Suppose that the SBN generates classification output $\hat{p}$ for a given input $\hat{x}$. The input to the SBN is obtained by applying bilateral mixup augmentation to two samples ($x_c, y_c)$ and ($x_r, y_r)$ generated by the uniform sampler and the re-balanced sampler, respectively
\begin{align}
    \lambda &\sim Beta(\alpha, \alpha), \label{eq:lambda}\\
    \left( \begin{matrix}\hat{x} \\ \hat{y} \end{matrix} \right) &= \lambda \left( \begin{matrix}{x}_c \\ {y}_c \end{matrix} \right) + (1-\lambda) \left( \begin{matrix}{x}_r \\ {y}_r \end{matrix} \right). \label{eq:mixup_input_c} 
\end{align}
 The resulting samples ($\hat{x}, \hat{y}$) are used to train the SBN through the loss function $\mathcal{L}_{total} = \mathcal{L}(\hat{p}, \hat{y})$. This method is referred to as {\it SBN-Mix}.

\input{table/cifarLT}

\section{Experiments}
\label{sec:expriments}
In this section, we evaluate the proposed method using four datasets for long-tailed visual recognition task: CIFAR-LT-10, CIFAR-LT-100, ImageNet-LT, and iNaturalist 2018. We also present an ablation study to evaluate the contribution of the components in the DBN-Mix.

\begin{table}[t]
    \centering
    \begin{flushleft}
    \begin{adjustbox}{width=0.70\columnwidth,center}
    \begin{tabular}{c|c|ccc}
        \Xhline{1.5pt}
        Method &  All & Many & Medium & Few\\
        \hline
        CE\;\; & 41.6 & 64.0 & 33.9 & 5.8 \\
        CE$^\S$ & 44.6 & -& -& -\\
        LDAM-DRW$^\S$ & 48.8 & - & -& -\\
        LWS\;\; & 47.7 & 57.1 & 45.2 & 29.3 \\ 
        LWS$^\S$ & 52.0 & 62.9& 49.8& 31.6\\
        MiSLAS$^\S$ & 52.7 & 61.7 & 51.3 & 35.8 \\
        GCL & 54.9 & - & - & - \\
        RIDE & 55.4 & 66.2 & 52.3 & 36.5\\
        RIDE + CMO & 56.2 & 66.4 & 53.9 & 35.6\\
        \hline
        DBN-Mix\;\; &  56.2 & 66.4 & \bf{54.9} & 33.7 \\ 
        DBN-Mix$^\S$& \bf{56.6} & \bf{67.9} & 53.2 & \bf{38.0}\\ 
        \Xhline{1.5pt}
    \end{tabular}
    \end{adjustbox}
    \end{flushleft}
    \caption{Top-1 test accuracy (\%) evaluated on ImageNet-LT. The results denoted by `$\S$' are obtained using a longer training schedule ($2\times$) following the settings in \cite{zhong2021mislas}.}
    \label{table:imagenet-LT}
\end{table}

\begin{table}[t]
    \centering
    \begin{flushright}
    \begin{adjustbox}{width=0.70\columnwidth,center}
    \begin{tabular}{c|c|ccc}
        \Xhline{1.5pt}
        Method &  All & Many & Medium & Few\\
        \hline
        CE & 61.7 & 72.2 & 63.0  & 57.2 \\ 
        LDAM-DRW & 68.0 & - & - & -\\
        BBN &  69.3 & 49.4 & 70.8 & 65.3 \\
        Remix & 70.5 & -& -& -\\
        cRT & 70.2 & \bf{74.2}& 71.1& 68.2\\
        LWS & 70.9 & 72.8& 71.6& 69.8\\
        MiSLAS & 71.6 & 73.2& 72.4& 70.4 \\
        GCL & 72.0 & - & - & - \\
        RIDE  & 72.6  & 70.9 & 72.4 & 73.1 \\
        RIDE + CMO & 72.8 & 68.7 & 72.6 & 73.1 \\
        \hline
        DBN-Mix  &  \bf{74.7} & 73.0 & \bf{75.6} & \bf{74.7} \\
        \Xhline{1.5pt}
    \end{tabular}
    \end{adjustbox}
    \end{flushright}
    \caption{Top-1 test accuracy (\%) on iNaturalist 2018. The performance of other methods is taken from the results reported in \cite{zhong2021mislas}.} 
    \label{table:iNaturalist18}
\end{table}
\subsection{Long-Tailed Recognition Datasets.}

\subsubsection{Long-tailed CIFAR} 
Long-tailed versions of the CIFAR datasets are artificially generated based on the original CIFAR-10 and CIFAR-100 datasets \cite{cui2019class}.  The degree of class imbalance in these datasets were specified by the imbalance ratio $\mu=\frac{\max_i{N_i}}{\min_j{N_j}}$, where $N_k$ be the number of training samples in the $k$-th class. In our experiments, we tried several imbalance ratios from $\mu\in\{10, 20, 50, 100, 200\}$. These imbalanced datasets are denoted CIFAR-LT-10 ($\mu$) and CIFAR-LT-100 ($\mu$), where $\mu$ represents the imbalance ratio. 

\subsubsection{Long-tailed ImageNet} The original ImageNet \cite{deng2009imagenet} is one of the largest image recognition datasets, which contains 1,280K training images and 50K test images with 1,000 categories. Following \cite{liu2019large}, we built a long-tailed version of the ImageNet dataset, which contains 115.8K training images. With this modification, the largest class size becomes 1,280 and the smallest size becomes 5.

\subsubsection{iNaturalist 2018} The iNaturalist 2018 dataset \cite{van2018inaturalist} is a large real-world dataset that exhibits a naturally imbalanced class distribution. This dataset contains 437.5K training images with 8,142 categories. Any modification was not applied to adjust the imbalance ratio. 

\subsection{Evaluation Metrics.}
The top-1 accuracy metric was used to evaluate the performance of the training methods. Test and validation sets have balanced sample sizes for all the classes. 
Following \cite{liu2019large}, we group the data samples into {\it Many}, {\it Medium} and {\it Few} classes based on their size, where {\it Many} denotes classes with more than 100 samples, {\it Medium} denotes classes with 20 to 100 samples, and {\it Few} denotes classes with less than 20 samples.

\subsection{Candidate Methods.}
Our method was compared to the following baseline algorithms: standard cross-entropy training (CE), Focal loss \cite{lin2017focal}, mixup \cite{Zhang2018mixup}, LDAM-DRW \cite{cao2019ldam}, M2m \cite{kim2020m2m}, Remix \cite{chou2020remix}, cRT \cite{kang2020decouple}, LWS \cite{kang2020decouple}, 
BBN \cite{zhou2020bbn}, Meta-weight net \cite{shu2019metaweight}, Meta-class weight \cite{muhammad2020metaclass} with focal loss,
MetaSAug \cite{li2021metasaug} with LDAM, Balanced Softmax \cite{ren2020balancedsoftmax}, MiSLAS \cite{zhong2021mislas}, PaCo \cite{cui2021parametric}, GCL \cite{li2022gcl}, RIDE (four experts) \cite{wang2021ride}, and RIDE (three experts) with CMO \cite{park2022cmo}. Unless otherwise noted, we used four expert configuration for RIDE.

\subsection{Implementation Details}
For both CIFAR-LT-10 and CIFAR-LT-100 datasets, we followed the setup in \cite{cao2019ldam}. We trained ResNet-32 \cite{he2016deep} with a batch size of 128 for 200 epochs. For ImageNet-LT, ResNet-50 \cite{he2016deep} was used with a batch size of 256. In the previous works, two different learning rate schedules were used to train the existing methods on ImageNet-LT dataset. For the {\it standard schedule}, the initial learning rate decayed by the factor of 0.1 at 60 and 80 epochs, and for the {\it extended schedule}, the learning rate decayed  at 120 and 160 epochs. In our experiments, we presented the results obtained using both schedules. For iNaturalist 2018, ResNet-50 \cite{he2016deep} was also used with a batch size of 256. We used the initial learning rate of 0.1 and decayed the learning rate at 120 and 160 epochs by 0.1. For all experiments, we used a stochastic gradient descent (SGD) optimizer with a momentum of 0.9 and set $\gamma=\infty$ for the re-balanced sampler, which led to class-balanced sampling. More detailed training setups are presented in Appendix A.

\subsection{Experimental Results}
\subsubsection{CIFAR-LT.}
Table \ref{table:cifar-LT} presents the classification accuracy of the proposed DBN-Mix method, compared with the existing methods on the CIFAR-LT-10 and CIFAR-LT-100 datasets. The proposed method outperforms existing methods by significant margins for all imbalance ratio configurations. In particular, for CIFAR-LT-10 (100) (i.e., an imbalance ratio of 100), DBN-Mix achieves a 1.37\% better performance than MiSLAS. Despite these performance improvements, DBN-Mix does not require two-stage training like MiSLAS. The performance gain of DBN-Mix is even higher on CIFAR-LT-100 (100).  DBN-Mix outperforms MiSLAS by 4.04\%. The proposed method outperforms the latest state-of-the-art  RIDE (three experts) + CMO by 1.04\%. Though the proposed method is based on two branch networks, it outperforms multiple expert networks with more expert branches. 


\subsubsection{ImageNet-LT.}
Table \ref{table:imagenet-LT} presents the top-1 accuracies of several methods evaluated on the ImageNet-LT dataset.
We observe that the proposed DBN-Mix also outperforms the existing methods. In the standard training schedule, DBN-Mix achieves 0.8\% better performance than RIDE. For an extended training schedule, DBN-Mix achieves a performance gain of 3.9\% over MiSLAS and 4.6\% over LWS. DBN-Mix improves the performance for both learning schedules, which shows that our method achieves a consistent performance improvement on long-tailed recognition.

\subsubsection{iNaturalist 2018.}
In Table \ref{table:iNaturalist18}, the performance of DBN-Mix is evaluated on iNaturalist 2018 dataset. DBN-Mix achieves the best classification accuracy among competitors. Our method achieves a 2.1\% performance gain over RIDE and 1.9\% performance gain over RIDE (3 experts) + CMO \cite{park2022cmo}, the CutMix-based state-of-the-art method.  
In particular, the performance gain of DBN-Mix over other methods is significant for the ``few" category, which shows that the proposed ideas are effective in recognizing the minority class samples.

\begin{table}[t]
    \setlength{\tabcolsep}{3pt}
    \centering
        \begin{flushleft}
        \begin{adjustbox}{width=1.0\columnwidth,center}
        \begin{tabular}{c|cc|ccc}
            \Xhline{1.5pt}
            \multirow{2}{*}{Method} & \multicolumn{2}{c|}{Main Components} & \multicolumn{3}{c}{Imbalance Ratio}\\
            \cline{2-6}
            & Bilateral Mixup & Temperature Scaling  & 100 & 50 & 10\\
            \hline
            Vanilla single-branch net. &  & & 38.46 & 44.02 & 55.73 \\ 
            \hline
            Single-branch net. + mixup \cite{Zhang2018mixup} & & &  39.54 & 44.99 & 58.02 \\
            \hline
            \multirow{3}{*}{Single-branch net. }
            & & $\checkmark$ & 41.25 & 45.64 & 58.36 \\ 
            & $\checkmark$ & & 44.10 & 49.73 & 61.98 \\ 
            & $\checkmark$ & $\checkmark$ & 45.07 & 50.39 & 62.37 \\ 
            \hline
            \multirow{4}{*}{Dual-branch net. } & & & 40.99  & 46.93 & 60.82 \\
            & & $\checkmark$ & 44.08 & 50.64 & 62.98 \\ 
            & $\checkmark$ &  & 46.61 & 51.42 & 63.59 \\ 
            & $\checkmark$ & $\checkmark$  & \bf{51.04} & \bf{54.93} & \bf{64.98}\\  
            \Xhline{1.5pt}
        \end{tabular}
        \end{adjustbox}
        \end{flushleft}
        \caption{Ablation study conducted on CIFAR-LT-100.}
        \label{table:ablation cifar100}
\end{table}

\subsection{Ablation Study}
\subsubsection{Contributions of Key Ideas.}
\label{sec:contribution of key ideas}
In Table \ref{table:ablation cifar100}, we analyze the impact of the two main ideas on overall performance: 1) bilateral mixup augmentation and 2) class-wise temperature scaling. The DBN structure used in BBN \cite{zhou2020bbn} was selected as another baseline. The vanilla SBN trained with cross-entropy loss is also selected as the baseline. 

Bilateral mixup offers a performance gain of 5.62\% over the DBN baseline on CIFAR-LT-100 (100). Temperature scaling offers a performance gain of 3.09\%. A considerable performance improvement of 10.05\% is achieved when bilateral mixup and temperature scaling are applied together. The performance gain of the proposed method increases as the imbalance ratio increases, which implies that the proposed DBN-Mix can handle severely imbalanced class distributions better than the other methods.

When we apply the bilateral mixup augmentation to the SBN baseline, classification accuracy improves by 5.64\%. We compare this result to the case where conventional mixup augmentation \cite{Zhang2018mixup} is applied to SBN. The performance gain achieved by the conventional mixup is only 1.08\%, demonstrating the superiority of the proposed bilateral mixup augmentation. Temperature scaling does not offer a large gain (i.e., 2.79\%) for SBN without bilateral mixup. However, when bilateral mixup augmentation and temperature scaling are used together, the performance gain increases dramatically to 6.61\%.

\begin{table}[t]
    \centering
        \begin{adjustbox}{width=0.95\columnwidth,center}
        \begin{tabular}{c|c|c|ccc}
            \Xhline{1.5pt}
            Method & Branch & All & Many & Medium & Few \\
            \hline
             & Re-balancing branch & 35.30 & 40.41 & 46.61 & 16.13 \\
             BBN & Conventional branch & 38.52 & 65.78 & 37.10 & 8.37 \\
             & Final & 42.95 & 65.86 & 45.38 & 13.40 \\
            \hline
             & Re-balancing branch & 47.16 & 56.37 & 54.18 & 28.23\\
           DBN-Mix  & Conventional branch & 48.03 & \bf{66.55} & 50.08 & 24.06\\
            & Final  & \bf{51.04} & 66.03 & \bf{55.49} & \bf{28.36} \\
            \Xhline{1.5pt}
        \end{tabular}
        \end{adjustbox}
        \caption{Performance of DBN-Mix evaluated at different branches and for different class groups. CIFAR-LT-100 (100) dataset was used for evaluation. } 
        \label{table:analysis branches}
\end{table}

\subsubsection{Performance Versus Hyperparameters.}
Additional ablation studies for investigating the impact of  several hyperparameters on performance are provided in Appendix B. We provide the performance of DBN-Mix as a function of the hyperparameters $\eta$, $\epsilon$, $\alpha$, and $\gamma$ evaluated on CIFAR-LT-10 (100) and CIFAR-LT-100 (100) datasets. From the experimental results in Appendix B, we set $\gamma=\infty$ that offers the best performance on both CIFAR-LT-10 and CIFAR-LT-100 datasets, respectively. We use the same hyperparameter value $\gamma=\infty$ for both the ImageNet-LT and iNaturalist 2018 datasets.



\subsection{Analysis and Discussion}
\subsubsection{Performance Evaluated at Different Branches and for Different Class Groups.}
Table \ref{table:analysis branches} reports a thorough analysis of the top-1 accuracy evaluated at different branches ({\it Re-balancing branch, Conventional branch, and Final output}) and for three class groups ({\it Many}, {\it Medium}, and {\it Few}). We measured the output accuracy for each branch.  {\it Many} denotes the set of  majority class samples and {\it Few} denotes the set of minority class samples. 

DBN-Mix offers significant performance improvements over the BBN baseline at both points. A performance gain of 11.86\% was achieved for the re-balancing branch and a performance gain of 9.51\% was achieved for the conventional learning branch. We also observe that for both the {\it Medium} and {\it Few} groups, DBN-Mix achieves significant performance gains. DBN-Mix maintains strong performance at all points for the {\it Many} group.

\begin{table}[t]
    \centering
    \begin{adjustbox}{width=0.95\columnwidth,center}
    \begin{tabular}{c|c|c|c}
        \Xhline{1.5pt}
        \multirow{2}{*}{Method} & \multicolumn{2}{c|}{Representation learning} & \multirow{2}{*}{Accuracy} \\
        \cline{2-3}
         & Bilateral Mixup & Temperature Scaling & \\
        \hline
        \multirow{4}{*}{Dual-branch net. }& & & 38.78 \\
        & & $\checkmark$ &  40.42 \\
        & $\checkmark$ & & 45.57 \\
        & $\checkmark$ & $\checkmark$ & \bf{46.25} \\
        \Xhline{1.5pt}
    \end{tabular}
    \end{adjustbox}
    \caption{Analysis of representation learning performance. CIFAR-LT-100 (100) dataset was used for evaluation.}
    \label{table:analysis representation}
\end{table}

\begin{table}[t]
    \centering
    \begin{adjustbox}{width=0.95\columnwidth,center}
    \begin{tabular}{c|c|c|c}
        \Xhline{1.5pt}
        \multirow{2}{*}{Method} & \multicolumn{2}{c|}{Classifier learning} & \multirow{2}{*}{Accuracy} \\
        \cline{2-3}
         & Bilateral Mixup & Temperature Scaling & \\
        \hline
        \hline
        \multirow{4}{*}{Dual-branch net. }& & & 43.23 \\
        & & $\checkmark$ & 45.86 \\
        & $\checkmark$ & & 42.67 \\
        & $\checkmark$ & $\checkmark$ & \bf{46.52} \\
        \Xhline{1.5pt}
    \end{tabular}
    \end{adjustbox}
    \caption{Analysis of classifier learning performance of DBN-Mix. CIFAR-LT-100 (100) dataset was used for evaluation.}
    \label{table:analysis classifier}
\end{table}

\subsubsection{Representation Learning Performance versus Classifier Learning Performance.}
\label{sec:representation_and_classifier}
A study published in \cite{kang2020decouple} showed that evaluating the performance of a model separately in terms of representation learning (RL) performance and classifier learning (CL) performance provides useful insights into understanding the behavior of its key components. 
Tables  \ref{table:analysis representation} and \ref{table:analysis classifier} show the RL and CL performances of DBN-Mix achieved by adding each idea individually. 
For RL performance, a classifier trained with cRT \cite{kang2020decouple} at the second stage was used. To evaluate the CL performance, a backbone network trained with conventional mixup augmentation \cite{Zhang2018mixup} for 200 epochs was used.
The bilateral mixup gives a significant performance improvement of 6.79\% in terms of the RL performance. In contrast, the bilateral mixup only marginally improves CL performance.  It can be seen that combining samples from two different samplers in the bilateral mixup serves to alleviate overfitting for the minority classes and thus improve the RL. We also see that the temperature scaling improves the RL performance by 1.64\% and the CL performance by 2.63\%. This shows that compensating for bias caused by the class imbalance improves both RL and CL performance.

\section{Conclusion}
In this paper, we studied the problem of training a DNN-based classification model using a dataset with a long-tailed class distribution. We proposed the bilateral mixup augmentation method to prevent the re-balancing branch of the DBN from degrading representation learning. The bilateral mixup achieved this goal simply by training with a convex combination of minority and majority training samples. We also proposed class conditional temperature scaling to compensate for the bias caused by class imbalance.  Our experiments on several long-tailed visual recognition datasets confirmed that the proposed DBN-Mix outperformed the DBN baseline and achieved state-of-the-art performance on various benchmarks.

\bibliography{aaai23.bib}

\clearpage

\appendix

\section{Appendix}
\subsection{A. Detailed Experimental Setup}
In this section, we provide the detailed experimental setups. Table \ref{table:training setup} presents the experimental setup used in our experiments.

\subsubsection{Long-tailed CIFAR} For both CIFAR-LT-10 and CIFAR-LT-100 datasets, ResNet-32 \cite{he2016deep} was trained with a batch size of 128.  We used a stochastic gradient descent (SGD) optimizer with a momentum of 0.9 and weight decay of $2\times10^{-4}$. Weight updates were performed over 200 epochs. Following the learning rate scheduling in \cite{cao2019ldam}, we set the initial learning rate to 0.1 and decayed the learning rate by a factor of 0.1 at 120 and 160 epochs, respectively. For the re-balanced sampler, we set $\gamma=\infty$, which led to class-balanced sampling. For temperature scaling, we chose $\eta=3, \epsilon=0.6$ for CIFAR-LT-10 and $\eta=7, \epsilon=0.6$ for CIFAR-LT-100. We applied standard data augmentation methods including horizontal flipping and random cropping \cite{he2016deep}. 

\subsubsection{Long-tailed ImageNet} In ImageNet-LT, ResNet-50 \cite{he2016deep} was used with a batch size of 256. Regarding SGD, the momentum, initial learning rate, and weight decay were set to 0.9, 0.2, and $2\times10^{-4}$, respectively. In the previous works, two different learning rate schedules were used to train the existing methods. For the {\it standard schedule}, the initial learning rate decayed by the factor of 0.1 at 60 and 80 epochs, and for the {\it extended schedule}, the learning rate decayed  at 120 and 160 epochs. In our experiments, we presented the results obtained using both schedules. The parameters for temperature scaling were set to $\eta=9, \epsilon=0.4$ and $\eta=9, \epsilon=0.2$ for the standard schedule and extended schedule, respectively. As for the re-balanced sampler, we set $\gamma=\infty$. Following the previous study \cite{he2016deep}, we applied horizontal flipping, resizing to 256$\times$256, and random cropping to 224$\times$224 for data augmentation. In the inference step, the 224$\times$224 patch was cropped from the center of the image and used as an input sample. 

\subsubsection{iNaturalist 2018} In iNaturalist 2018, ResNet-50 \cite{he2016deep} was also used. The momentum, initial learning rate, and weight decay were set to 0.9, 0.1, and $1\times10^{-4}$, respectively. The learning rate decayed by the factor of 0.1 at 120 and 160 epochs. We used the hyperparameter setup, $\gamma=\infty, \eta=10$, and $\epsilon=2\times10^{-2}$. We applied the same data augmentation used for ImageNet-LT.

\subsection{B. Performance Versus Hyperparameters}

\begin{figure*}[th]
    \begin{subfigure}[b]{0.32\textwidth}
        \centering
        \includegraphics[width=0.98\columnwidth]{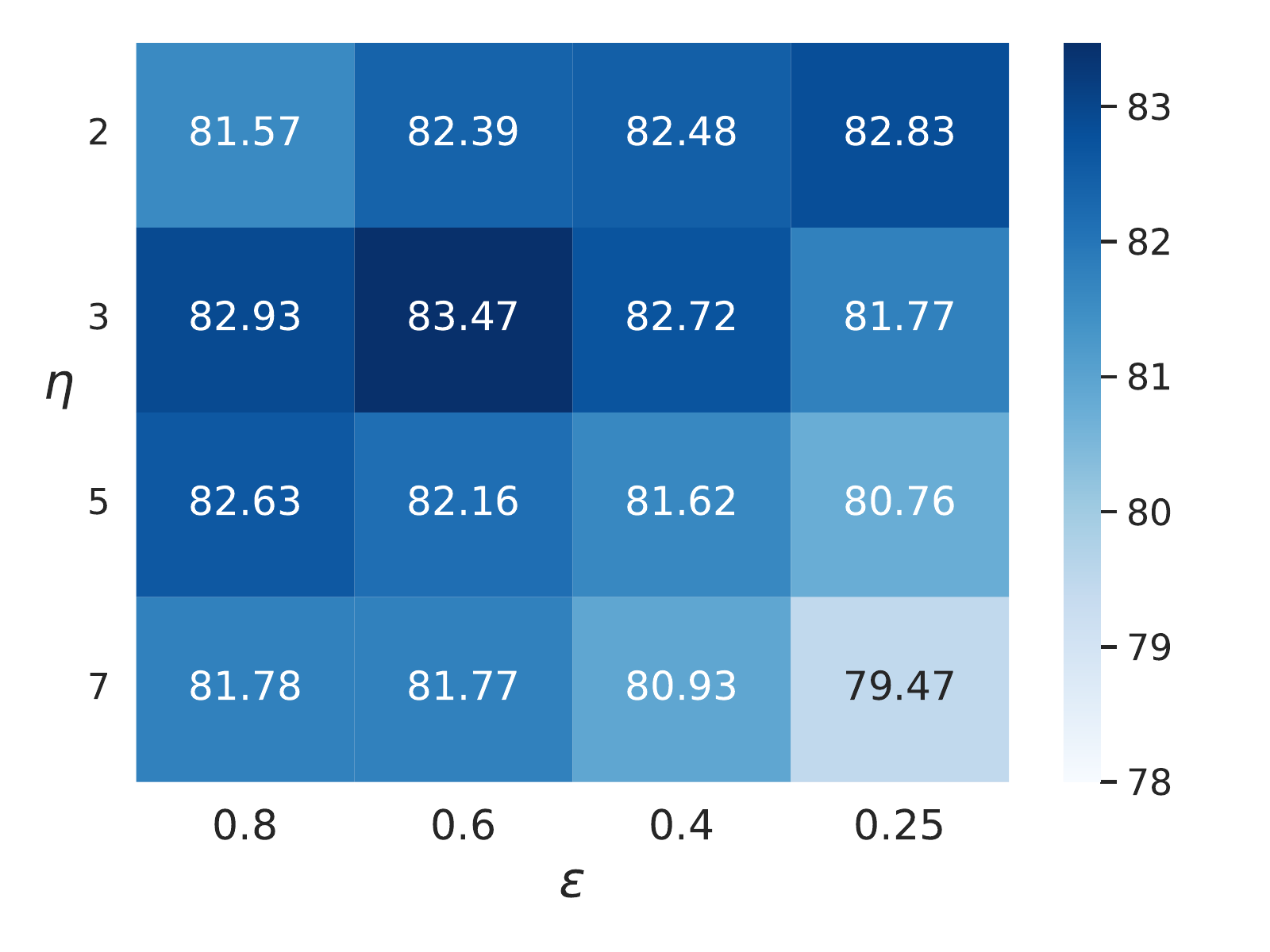}
        \caption[]{}
        \label{fig:ablation cifar10 temperature}
    \end{subfigure}
    \begin{subfigure}[b]{0.32\textwidth}
        \centering
        \includegraphics[width=0.98\columnwidth]{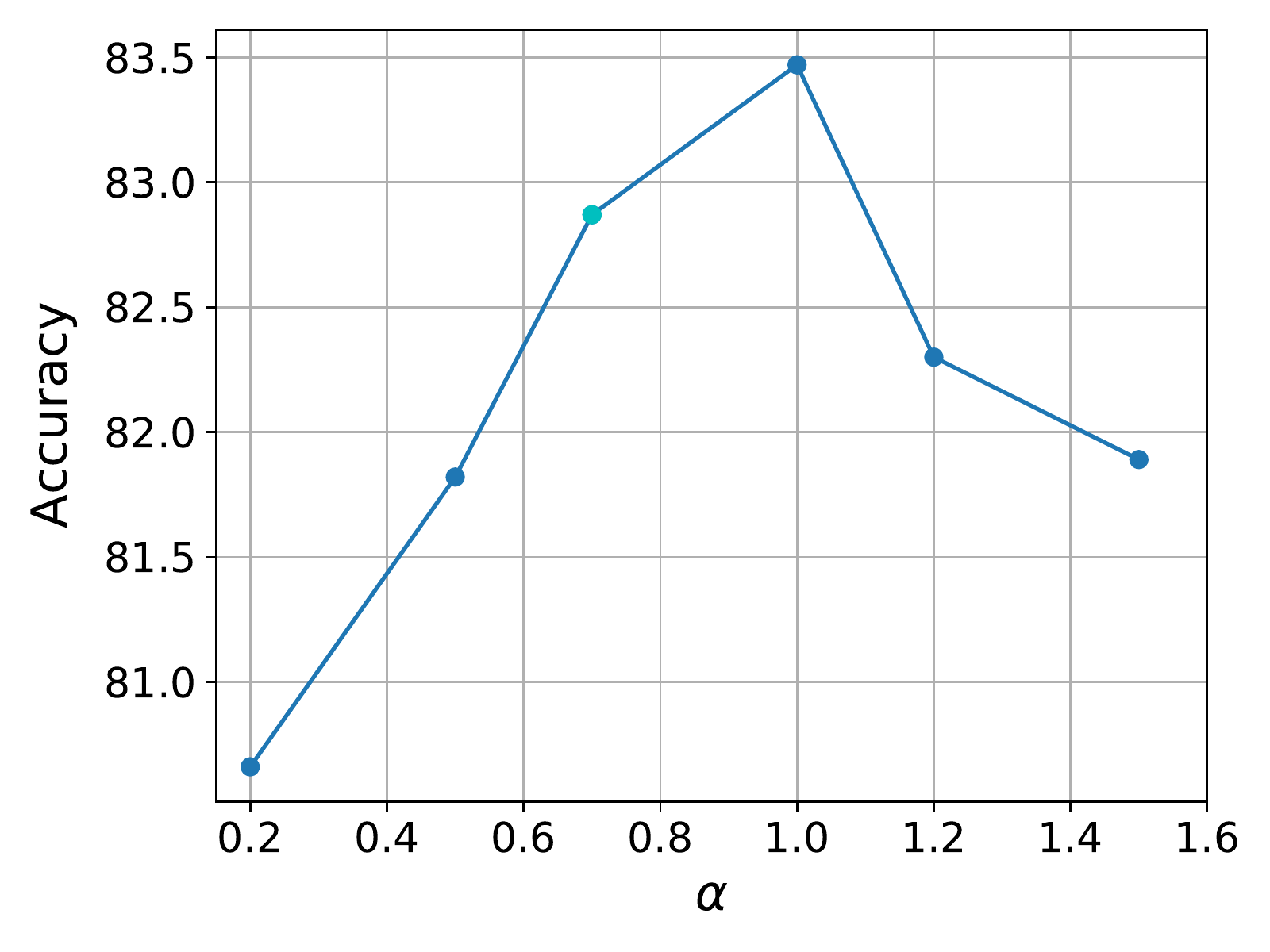} 
        \caption[]{}
        \label{fig:ablation cifar10 alpha}
    \end{subfigure}
    \begin{subfigure}[b]{0.32\textwidth}  
        \centering 
        \includegraphics[width=1.0\columnwidth]{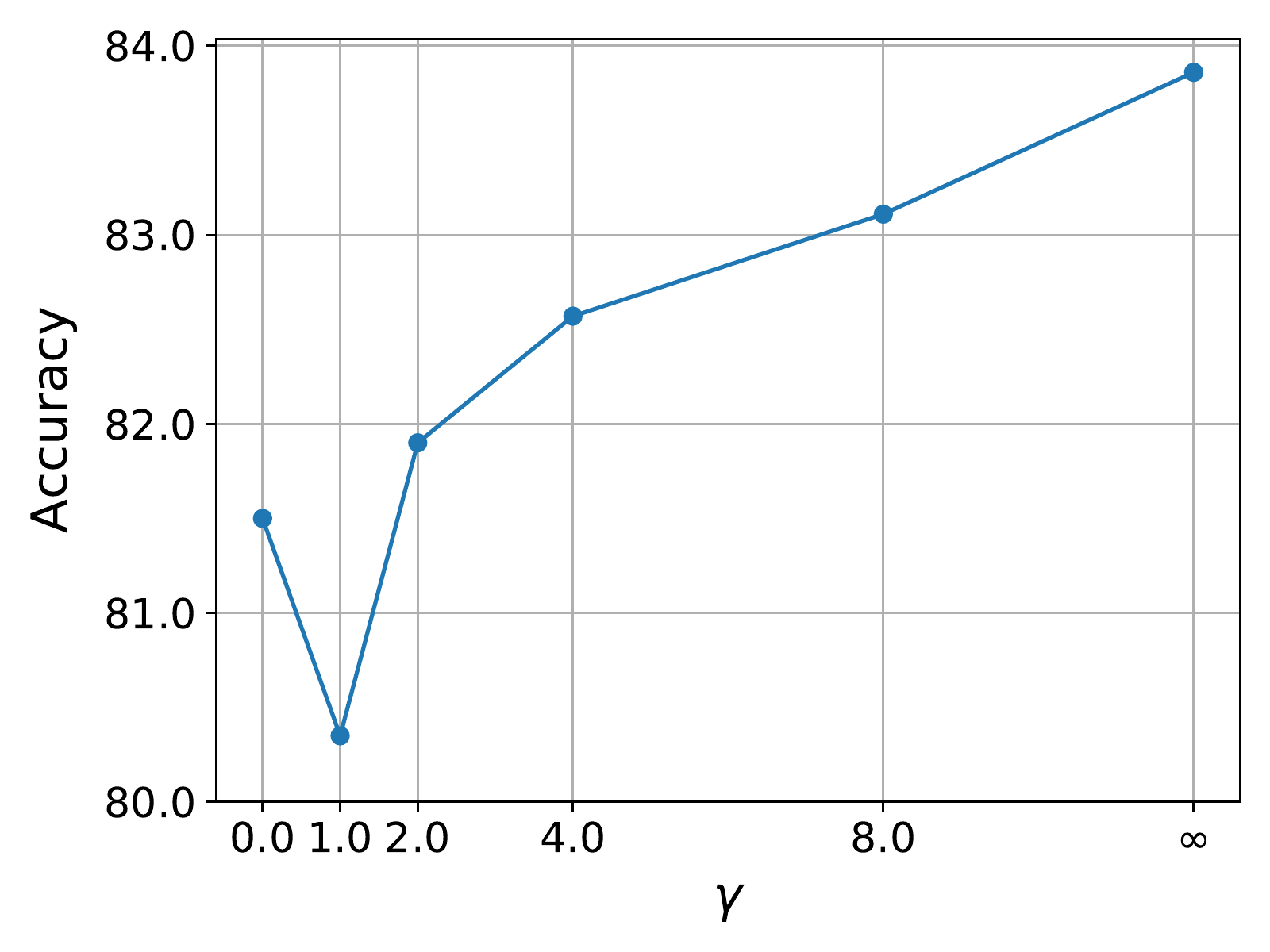} 
        \caption[]{}
        \label{fig:ablation cifar10 gamma}
    \end{subfigure}
    \caption{Performance versus hyperparameters: (a) $\eta$ and $\epsilon$ for temperature scaling, (b) $\alpha$ for bilateral mixup augmentation, and (c) $\gamma$ for re-balanced sampler. CIFAR-LT-10 (100) dataset was used for evaluation.} 
    \captionsetup{singlelinecheck = false}
    \label{fig:ablation cifar10 hyperparameters}
\end{figure*}

\begin{figure*}[th]
    \begin{subfigure}[b]{0.32\textwidth}
        \centering
        \includegraphics[width=0.98\columnwidth]{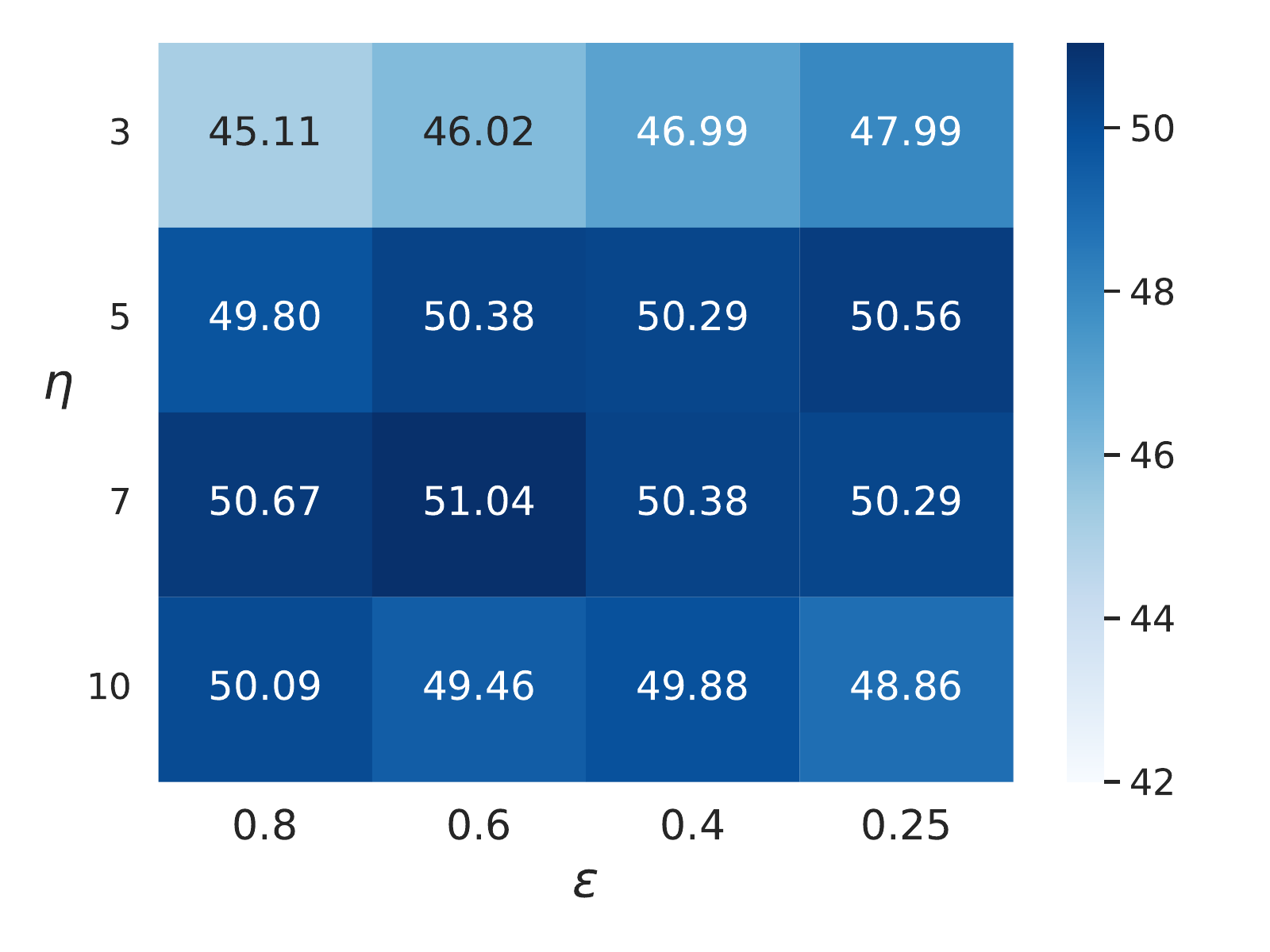}
        \caption[]{}
        \label{fig:ablation cifar100 temperature}
    \end{subfigure}
    \begin{subfigure}[b]{0.32\textwidth}
        \centering
        \includegraphics[width=0.98\columnwidth]{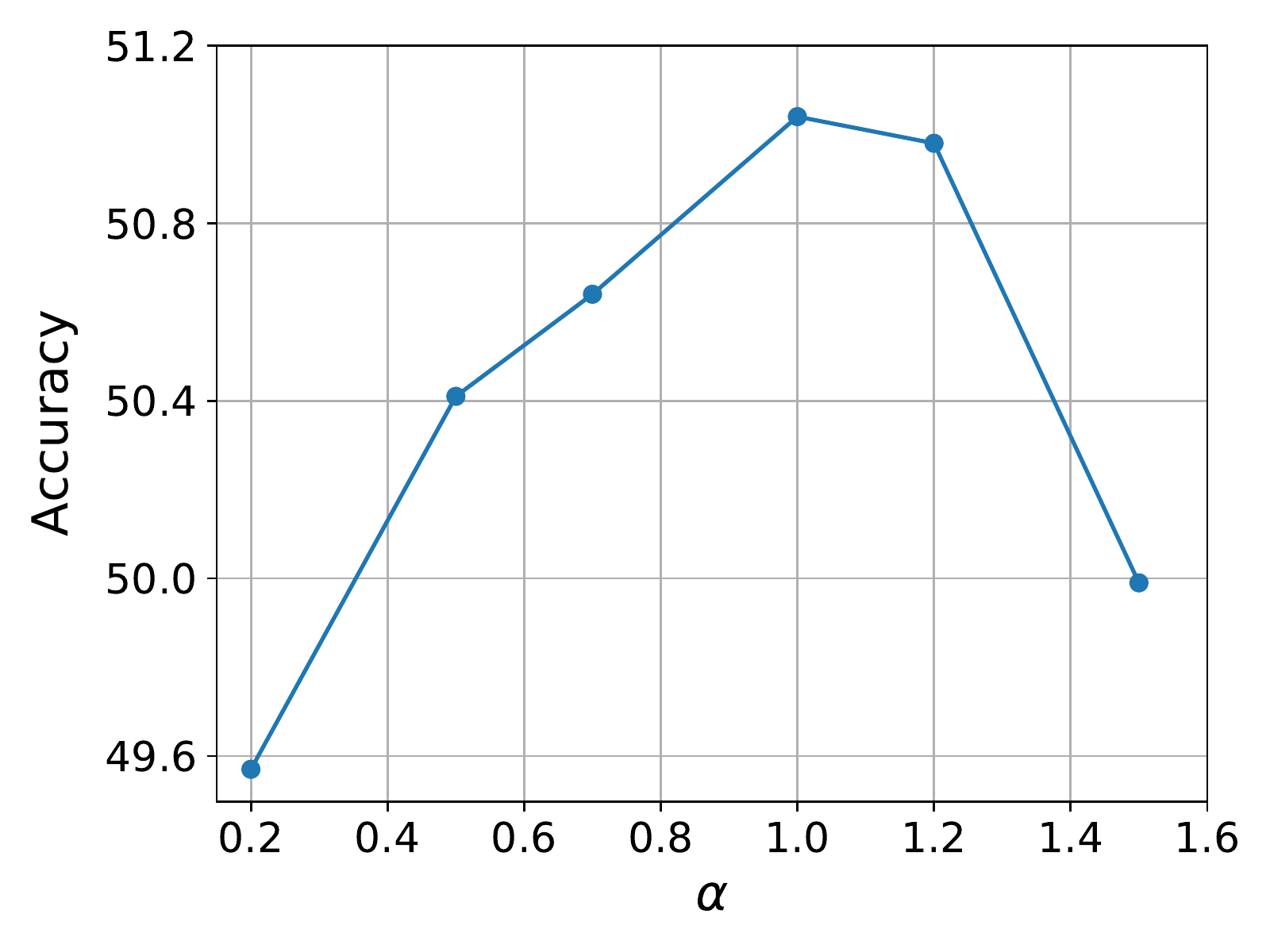} 
        \caption[]{}
        \label{fig:ablation cifar100 alpha}
    \end{subfigure}
    \begin{subfigure}[b]{0.32\textwidth}  
        \centering 
        \includegraphics[width=1.0\columnwidth]{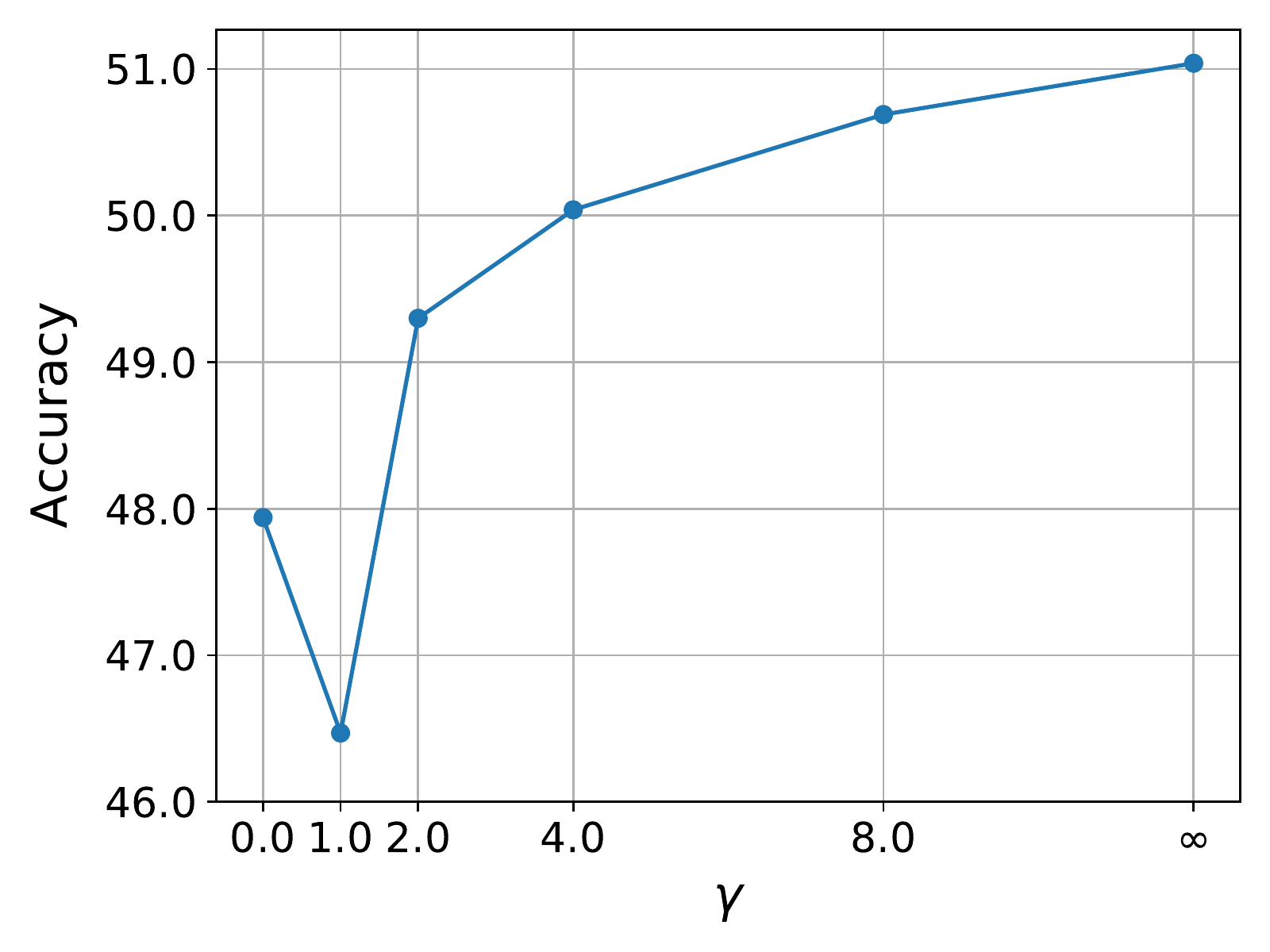} 
        \caption[]{}
        \label{fig:ablation cifar100 gamma}
    \end{subfigure}
    \caption{Performance versus hyperparameters: (a) $\eta$ and $\epsilon$ for temperature scaling, (b) $\alpha$ for bilateral mixup augmentation, and (c) $\gamma$ for re-balanced sampler. CIFAR-LT-100 (100) dataset was used for evaluation.} 
    \captionsetup{singlelinecheck = false}
    \label{fig:ablation cifar100 hyperparameters}
\end{figure*}

Fig. \ref{fig:ablation cifar10 hyperparameters} (a) and Fig. \ref{fig:ablation cifar100 hyperparameters} (a) present the performance of DBN-Mix as a function of the parameters $\eta$ and $\epsilon$ evaluated on CIFAR-LT-10 (100) and CIFAR-LT-100 (100), respectively. A larger $\epsilon$ or smaller $\eta$ enhances the effect of temperature scaling. Through extensive experiments, we find that the setup $\eta=3, \epsilon=0.6$ and $\eta=7, \epsilon=0.6$ offers the best performances for CIFAR-LT-10 and CIFAR-LT-100, respectively.

Fig. \ref{fig:ablation cifar10 hyperparameters} (b) and Fig. \ref{fig:ablation cifar100 hyperparameters} (b) present the performance with respect to $\alpha$, the parameter of the beta distribution used in the bilateral mixup. We tried different values of $\alpha$ from the set $\alpha\in\{0.2, 0.5, 0.7, 1.0, 1.2, 1.5\}$. The best performance is achieved at $\alpha=1.0$ for both the CIFAR-LT-10 and CIFAR-LT-100 datasets. The best performance is achieved at $\alpha=0.2$ for the ImageNet-LT and iNaturalist 2018 datasets.

Finally, Fig. \ref{fig:ablation cifar10 hyperparameters} (c) and Fig. \ref{fig:ablation cifar100 hyperparameters} (c) provide the performance of DBN-Mix as a function of $\gamma$ used in the re-balanced sampler. The re-balancing effect on the class distribution is stronger as $\gamma$ increases. The class-balanced sampling $\gamma=\infty$ offers the best performance.

\subsection{C. Training on Longer Schedule with Strong Augmentation Method}
Following the experimental setup used in \cite{cui2021parametric, park2022cmo}, 
we trained the model over 400 epochs with AutoAugment \cite{cubuk2019autoaugment} on CIFAR-LT-100. Table \ref{table:strong augment} shows that the proposed method outperforms the existing methods with strong augmentation by significant margins for all the imbalance ratios considered. In particular, for the imbalance ratio of 100, DBN-Mix outperforms PaCo by 2.3\% and Balanced Softmax with CMO by 2.6\%. A notable point is that an additional strong augmentation method \cite{cubuk2019autoaugment} can significantly improve our bilateral mixup even if our method is much simpler in terms of structure.

\begin{table}[th]
    \centering
        \begin{adjustbox}{width=1.0\columnwidth,center}
        \begin{tabular}{c|ccc}
            \Xhline{1.5pt}
            \multirow{2}{*}{Method} & \multicolumn{3}{c}{Imbalance Ratio} \\
            \cline{2-4}
            & 100 & 50 & 10\\
            \hline
            Balanced Softmax \cite{ren2020balancedsoftmax} & 50.8 & 54.2 & 63.0 \\
            PaCo \cite{cui2021parametric} & 52.0 & 56.0 & 64.2 \\
            Balanced Softmax + CMO \cite{park2022cmo} & 51.7 & 56.7 & 65.3 \\
            \hline
            DBN-Mix & \bf{54.3} & \bf{57.7} & \bf{66.4}\\
            \Xhline{1.5pt}
        \end{tabular}
        \end{adjustbox}
        
        \caption{Top-1 test accuracy (\%) on CIFAR-LT-100. Performance of DBN-Mix evaluated for the imbalance ratio from $\mu \in \{10, 50, 100\}$. Following the experimental setup used in \cite{park2022cmo}, each method trains the model for 400 epochs with AutoAugment \cite{cubuk2019autoaugment}.}
        \label{table:strong augment}
\end{table}

\subsection{D. T-SNE Visualization Evaluated at Different Branches}
\begin{figure*}[th]
    \centering
    \includegraphics[width=0.7\textwidth]{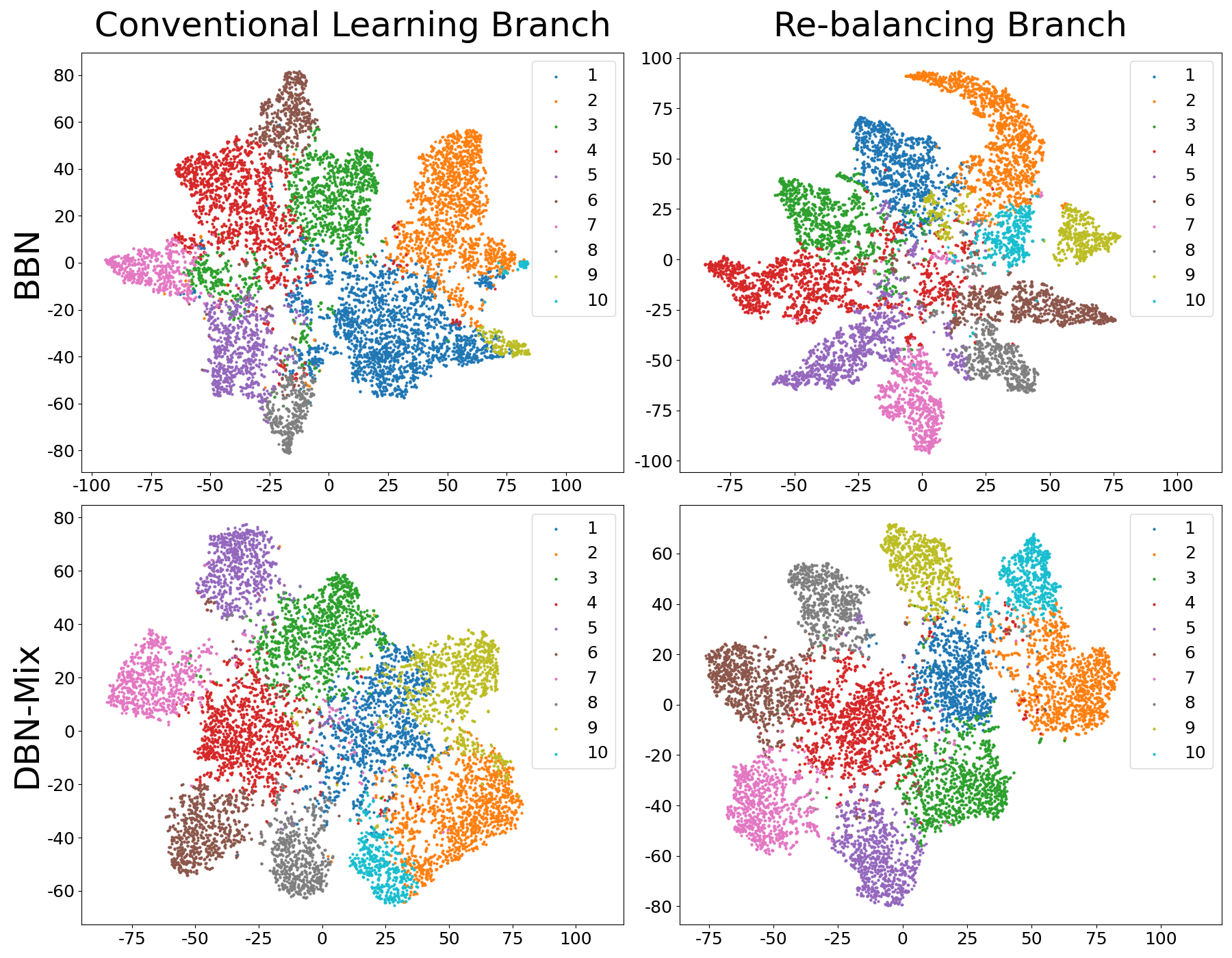}
    \caption[]{T-SNE \cite{maaten2008visualizing} illustrates the feature at the penultimate layer of the conventional learning branch and re-balancing branch. The feature of the conventional learning branch (first column) and the re-balancing branch (second column) in DBN are trained by DBN-Mix (first row) and BBN (second row) \cite{zhou2020bbn}, respectively.}
    \label{fig:T-SNE}
\end{figure*}

We visualize the feature vector of the penultimate layer of conventional learning branch and re-balancing branch using T-SNE \cite{maaten2008visualizing}. The T-SNE visualization method projects a feature vector onto a lower-dimensional embedding space. Following the experimental setup used in Table. 1, we train the DBN on CIFAR-LT-10 with an imbalance ratio of 100. Fig. \ref{fig:T-SNE} shows the feature of the conventional learning branch and the re-balancing branch in DBN trained by BBN and DBN-Mix, respectively. For the features trained by BBN, we can observe that the features of each class are entangled in both the conventional learning branch and the re-balancing branch, making it difficult to distinguish them from other classes. Contrary to these results, DBN-Mix obtains more clear decision boundaries and learns better representations for all classes.

\begin{table*}[th]
    \centering
        \begin{adjustbox}{width=0.95\textwidth,center}
        \begin{tabular}{c|cccc|cccc}
            \Xhline{1.5pt}
            \multirow{2}{*}{Dataset} & \multicolumn{4}{c|}{Common setting} & \multicolumn{4}{c}{Hyperparameters for DBN-Mix}\\
            \cline{2-9}
            & Batch Size & Initial Learning Rate & Weight Decay & Momentum & $\alpha$ & $\gamma$ & $\eta$ & $\epsilon$ \\
            
            \hline
            CIFAR-LT-10      & 128 & 0.1 & 2$\times10^{-4}$ & 0.9 & 1.0 & $\infty$& 3& 0.6\\
            CIFAR-LT-100     & 128 & 0.1 & 2$\times10^{-4}$ & 0.9& 1.0 & $\infty$& 7& 0.6\\
            ImageNet-LT (Standard) & 256 & 0.2 & 2$\times10^{-4}$ & 0.9& 0.2 & $\infty$& 9&0.4\\
            ImageNet-LT (Extended)     & 256 & 0.2 & 2$\times10^{-4}$ & 0.9& 0.2 & $\infty$& 9&0.2\\
            iNaturalist 2018 & 256 & 0.1 & 1$\times10^{-4}$ & 0.9& 0.2 & $\infty$ & 10& $2\times10^{-2}$\\
            
            \Xhline{1.5pt}
        \end{tabular}
        \end{adjustbox}
        
        \caption{Detailed experimental configuration for our experiments.}
        \label{table:training setup}
\end{table*}

\clearpage

\end{document}

%% file: table/cifarLT.tex
\begin{table*}[th]
\centering
  \begin{adjustbox}{width=0.9\textwidth,center}
  \begin{tabular}{c|ccccc|ccccc}
    
    \Xhline{1.5pt}
    Dataset & \multicolumn{5}{c|}{CIFAR-LT-10} & \multicolumn{5}{c}{CIFAR-LT-100}\\
    \hline
    Imbalance Ratio & 200 & 100 & 50 & 20 & 10 & 200 & 100 & 50 & 20 & 10 \\
    \hline
    Cross-entropy\;\;\; & 65.87 & 70.14 & 74.94 & 82.44 & 86.18& 34.70 & 38.46 & 44.02 & 51.06 & 55.73\\
    Focal loss \cite{lin2017focal} & 65.29 & 70.38 & 76.71 & 82.76 & 86.66 & 35.62 & 38.41 & 44.32 & 51.95 & 55.78  \\
    Mixup$\dagger$ \cite{Zhang2018mixup} & - & 73.06 & 77.82 & - & 87.10 & - & 39.54 & 44.99 & - & 58.02 \\
    LDAM-DRW \cite{cao2019ldam}&- & 77.03  & - & - & 88.16 & 38.45 & 42.89 & 47.97 & 52.99 & 58.78  \\
    M2m + LDAM \cite{kim2020m2m} & - & 79.10 & - & - & 87.50 & - & 43.50 & - & - & 57.60 \\
    Remix \cite{chou2020remix} & - & 79.76 & - & - & 89.02 & - & 46.77 & - & - & 61.23 \\
    BBN$\dagger$ \cite{zhou2020bbn}& -&  79.82 & 82.18 & - & 88.32 & - & 42.56 & 47.02 & - & 59.12\\
    Meta-weight net$\ddagger$ \cite{shu2019metaweight} & 67.20 & 73.57 & 79.10 & 84.45 & 87.55& 36.62 & 41.61 & 45.66 & 53.04 & 58.91  \\
    MCW$\ddagger$ + Focal \cite{muhammad2020metaclass} & 74.43 & 78.90 & 82.88 & 86.10 & 88.37 & 39.34 & 44.70 & 50.08 & 55.73 & 59.59 \\
    MetaSAug + LDAM \cite{li2021metasaug}& 77.35 & 80.66 & 84.34 &  88.10 & 89.68 & 43.09 & 48.01 & 52.27 & 57.53 & 61.28 \\
    MiSLAS \cite{zhong2021mislas} & - & 82.10 & 85.70 & - & 90.00  & - & 47.00 & 52.30 & - & 63.20 \\
    GCL \cite{li2022gcl} & 79.03 & 82.68 & 85.48 & - & -  & 44.88 & 48.71 & 53.55 & - & - \\
    RIDE \cite{wang2021ride} & - & - & - & - & - & - &  49.10 & - & - & -\\
    RIDE + CMO \cite{park2022cmo} & - & - & - & - & - & - &  50.00 & 53.00 & - & 60.20 \\
    \hline
    SBN-Mix & 69.87 & 76.33 & 81.04 & 86.91 & 89.84 & 40.30 & 45.07 & 50.39 & 57.28 & 62.37 \\
    DBN-Mix & \bf{79.58} & \bf{83.47} & \bf{86.82} & \bf{89.11} & \bf{90.87} & \bf{46.21} & \bf{51.04} & \bf{54.93} & \bf{61.07} & \bf{64.98}  \\

    \Xhline{1.5pt}
    \end{tabular}
  \end{adjustbox}
  \caption{Top-1 test accuracy (\%) evaluated on CIFAR-LT-10 and CIFAR-LT-100 with various imbalance ratios. The entries denoted by `$\dagger$' and `$\ddagger$' are taken from the results reported in \cite{zhou2020bbn} and \cite{muhammad2020metaclass}, respectively.} 
  \label{table:cifar-LT}
\end{table*}